\documentclass[letterpaper, 10 pt, conference]{ieeeconf}  % Comment this line out if you need a4paper

\IEEEoverridecommandlockouts                              % This command is only needed if 
\usepackage{textcomp}
\usepackage{stfloats}
\usepackage{url}
\usepackage{verbatim}
\usepackage{graphicx}
\usepackage{cite}
\usepackage{amsmath,amssymb,amsfonts}
\usepackage{algorithmic}
\usepackage{graphicx}
\usepackage{textcomp}
\usepackage{graphics} % for pdf, bitmapped graphics files
\usepackage{url}
\usepackage{bm}
\usepackage{caption}
\usepackage{graphicx}
\usepackage{cite}
\usepackage{multirow}
\usepackage{enumerate}
\usepackage{amsmath}
\usepackage{etoolbox}
\usepackage{placeins}
\usepackage{hyperref}
\usepackage{setspace} 
\usepackage{romannum}
\usepackage{wasysym}
\usepackage{subcaption}

\usepackage{colortbl}
\usepackage{booktabs}
\usepackage{multirow}
\let\labelindent\relax
\usepackage{enumitem}
\usepackage[normalem]{ulem}
\useunder{\uline}{\ul}{}
\usepackage{pifont}
\usepackage{wasysym}
\usepackage{amssymb}
\usepackage{pifont}% http://ctan.org/pkg/pifont
\usepackage{tikz}
\usepackage{xcolor}
\usepackage{etoolbox}
\usepackage{subcaption} 
\definecolor{Green}{RGB}{0, 255, 0}
\definecolor{Red}{RGB}{255, 0, 0}
\newcommand{\cmark}{\textcolor{Green}{\ding{51}}}%
\newcommand{\xmark}{\textcolor{Red}{\ding{55}}}%
\usepackage{siunitx}

\setlist[itemize]{leftmargin=*}

\title{\LARGE \bf
GraspClutter6D: A Large-scale Real-world Dataset \\ for Robust Perception and Grasping in Cluttered Scenes}

\author{Seunghyeok Back$^{1}$, Joosoon Lee$^{2}$, Kangmin Kim$^{2}$, Heeseon Rho$^{2}$, Geonhyup Lee$^{2}$, Raeyoung Kang$^{2}$, \\ Sangbeom Lee$^{2}$, Sangjun Noh$^{2}$, Youngjin Lee$^{2}$, Taeyeop Lee$^{3}$ and Kyoobin Lee$^{2\dagger}$
\thanks{$^{1}$ Korea Institute of Machinery \& Materials (KIMM)}  
\thanks{$^{2}$ Gwangju Institute of Science and Technology (GIST)}  
\thanks{$^{3}$ Korea Advanced Institute of Science and Technology (KAIST)} 
\thanks{S. Back was with GIST during initial research and is now with KIMM.} 
\thanks{$^{\dagger}$ Corresponding author: Kyoobin Lee {\tt\small kyoobinlee@gist.ac.kr}}  
}

\begin{document}

\maketitle
% \thispagestyle{empty}
% \pagestyle{empty}

% collected through multi-sensor robotic systems and crowdsourced annotation
%%%%%%%%%%%%%%%%%%%%%%%%%%%%%%%%%%%%%%%%%%%%%%%%%%%%%%%%%%%%%%%%%%%%%%%%%%%%%%%%
\begin{abstract}
Robust grasping in cluttered environments remains an open challenge in robotics. While benchmark datasets have significantly advanced deep learning methods, they mainly focus on simplistic scenes with light occlusion and insufficient diversity, limiting their applicability to practical scenarios. We present GraspClutter6D, a large-scale real-world grasping dataset featuring: (1) 1,000 highly cluttered scenes with dense arrangements (14.1 objects/scene, 62.6\% occlusion), (2) comprehensive coverage across 200 objects in 75 environment configurations (bins, shelves, and tables) captured using four RGB-D cameras from multiple viewpoints, and (3) rich annotations including 736K 6D object poses and 9.3B feasible robotic grasps for 52K RGB-D images. We benchmark state-of-the-art segmentation, object pose estimation, and grasp detection methods to provide key insights into challenges in cluttered environments. Additionally, we validate the dataset's effectiveness as a training resource, demonstrating that grasping networks trained on GraspClutter6D significantly outperform those trained on existing datasets in both simulation and real-world experiments. The dataset, toolkit, and annotation tools are publicly available on our project website: \url{https://sites.google.com/view/graspclutter6d}.
\end{abstract}
% \begin{IEEEkeywords}
% Data sets for robotic vision, Data sets for robot learning, Deep learning in grasping and manipulation.
% \end{IEEEkeywords}

\section{INTRODUCTION}

Grasping is one of the most fundamental yet challenging tasks in robotics, with applications spanning warehouses, manufacturing, and household assistance. Effective robotic grasping requires coordination of multiple visual perception capabilities: segmentation \cite{he2017mask, cheng2022masked, kirillov2023segment}, 6D object pose estimation \cite{xiang2018posecnn, he2021ffb6d, wen2024foundationpose}, and 6-DoF grasp detection \cite{fang2020graspnet, sundermeyer2021contact, fang2023anygrasp}. Recent significant progress has been driven by comprehensive datasets \cite{rennie2016dataset, zeng2017multi, hodan2017t, xiang2018posecnn, suchi2019easylabel, fang2020graspnet, gilles2023metagraspnetv2, hodan2024bop, fang2023anygrasp}, enabling deep learning to achieve better generalization. However, grasping in \textit{cluttered environments}—a common scenario in practical settings—remains challenging. In these environments, objects are densely packed in unknown poses under varied backgrounds, requiring sophisticated approaches to handle occlusion \cite{newbury2023deep, bauer2024challenges}. 

Although datasets have advanced this field, most existing benchmarks focus on structured, simplified scenes rather than cluttered environments. GraspNet-1Billion (GraspNet-1B) \cite{fang2020graspnet, fang2023robust}, a widely adopted benchmark, exhibits only modest complexity (avg. 8.9 objects per scene with 35.2\% occlusion) and restricts diversity to single tabletop scenes without background variation. Similarly, the recent HouseCat6D dataset \cite{jung2024housecat6d} offers greater object diversity (192 objects) but limited scene composition, containing only 41 scenes with minimal occlusion (23.5\%). This gap between simplified datasets and real-world complexity presents a significant challenge in developing robust robot manipulation systems.

\begin{figure}[t!]
    \centering
         \includegraphics[width=0.98\columnwidth]{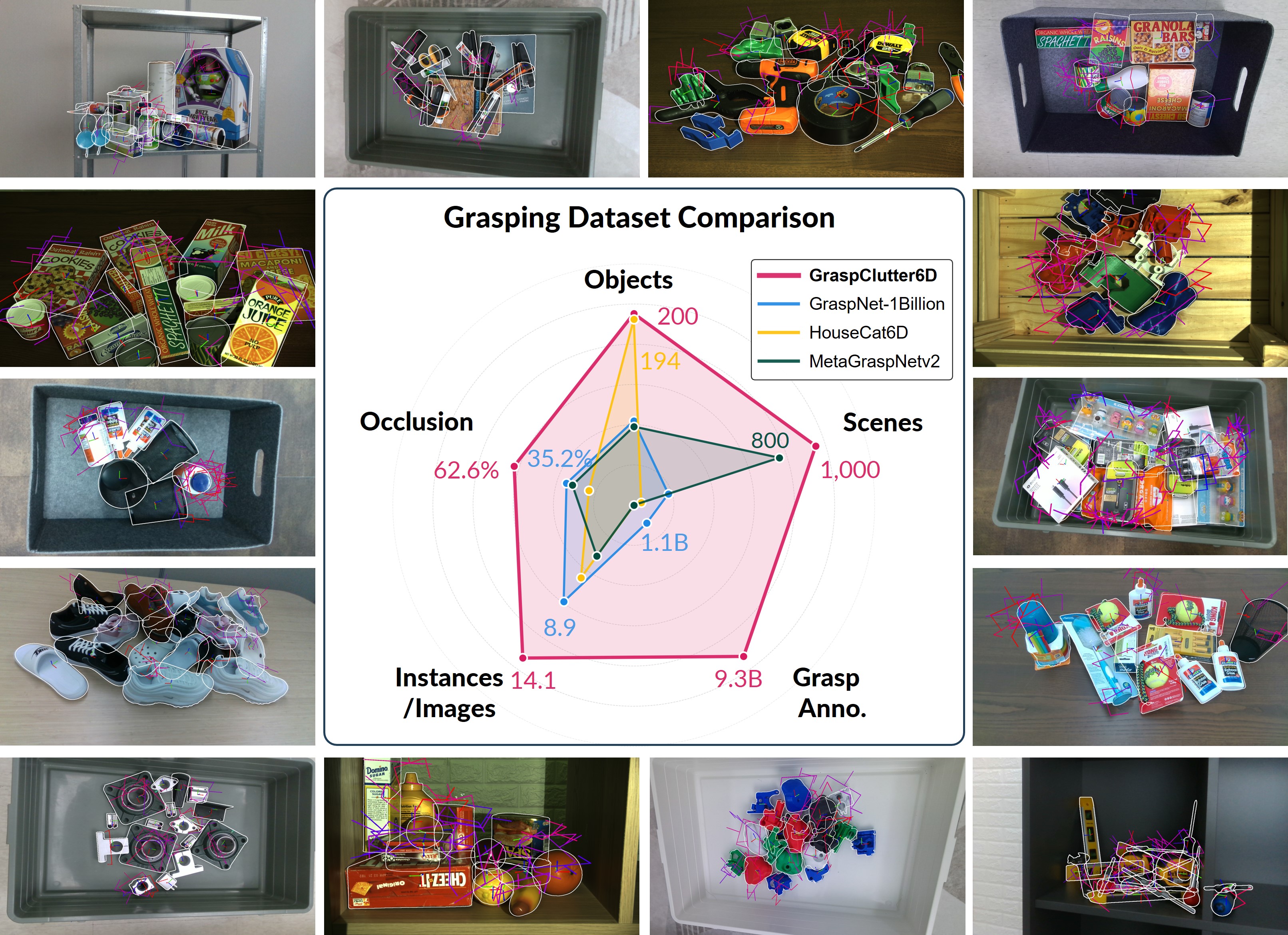}
    \caption{We present \textbf{GraspClutter6D}, a large-scale dataset with high complexity (avg. \textbf{14.1} instances, \textbf{62.6\%} occlusion), including \textbf{200} objects in \textbf{1,000} scenes from four RGB-D sensors. It offers \textbf{736K} 6D object poses and \textbf{9.3B} grasps. (25 grasps are visualized per image (avg. 178K grasps / image))}
    \label{fig:fig1}
\end{figure}

In this work, we present \textbf{GraspClutter6D}, a large-scale dataset for robotic grasping and perception (Fig. \ref{fig:fig1}), the most extensive real-world resource with diverse objects in highly cluttered environments. Through multi-sensor robotic capture and crowd-sourcing, we collected 1,000 highly cluttered scenes containing 9.3 billion grasp annotations and 736K instance annotations. Benchmark evaluations reveal that state-of-the-art grasping methods face significant challenges in our cluttered environments, while training on our dataset substantially improves grasping performance compared to existing datasets. We have released the complete dataset, annotation tools, and purchase links for all objects and furniture to facilitate research in robust robotic manipulation.

The contributions of this work are summarized as follows:
\begin{itemize}
\item \textbf{Real-world dense clutter}: 1,000 densely packed scenes with high complexity, averaging 14.1 instances with 62.6\% occlusion, totaling 52K RGB-D images.
\item \textbf{Diverse coverage}: Bin/shelf/table scenes with 200 objects captured using four cameras from multiple viewpoints.
\item \textbf{Extensive annotations}: 9.3 billion 6-DoF grasp poses with 736K 6D object poses and segmentation masks.
\item \textbf{Improved grasping and perception}: Models trained on our dataset demonstrate significantly better performance than those trained on existing benchmark datasets.
\item \textbf{Benchmark evaluation}: Assessment of state-of-the-art grasping and perception methods to provide baselines.

\end{itemize}

\begin{table*}[t!]
\vspace*{0.3cm}
\caption{\textbf{Comparison of real-world 6D pose estimation and grasping datasets.} GraspClutter6D provides highly complex scenes at scale in diverse environments. \textit{Visibility}: mean visible pixel ratio per instance; \textit{Occlusion}: percentage of instances with Visibility $\leq$ 0.95. $^*$OCID: planar grasps only; others: 6D grasps. (K$=$thousands, M$=$millions, B$=$billions)}
% GraspClutter6D provides \textbf{extensive coverage at scale} with 200 objects, 1,000 scenes, and 4 RGB-D sensors, featuring \textbf{highest complexity} (14.1 instances/image, 62.6\% occlusion) in bin, shelf, table environments. 
\large
\centering
\resizebox{2.0\columnwidth}{!}
{
\def\arraystretch{1.0}%  1 is the default, change whatever you need
\setlength\tabcolsep{1.0em}
\begin{tabular}{lcccrrccrrccl}
\hline
\multirow{2}{*}{\textbf{Dataset}} & \multicolumn{3}{c}{\textbf{Annotations}} & \multicolumn{5}{c}{\textbf{Scale}} & \multicolumn{4}{c}{\textbf{Scene Complexity}} \\ \cmidrule(lr){2-4}\cmidrule(lr){5-9}\cmidrule(l){10-13}
 & Segm. & 6D Pose & Grasp & Objects & Scenes & \begin{tabular}[c]{@{}c@{}}Sensor\\ Type(s)\end{tabular} & Grasps & Images & \begin{tabular}[c]{@{}r@{}}Instances\\ / Image $\bm{\uparrow}$\end{tabular} & \begin{tabular}[c]{@{}c@{}}Visibility\\ (\%) $\bm{\downarrow}$\end{tabular} & \begin{tabular}[c]{@{}c@{}}Occlusion\\ (\%) $\bm{\uparrow}$\end{tabular} & \begin{tabular}[l]{@{}l@{}}Environment\\ Type(s)\end{tabular} \\ \cmidrule(r){1-1}\cmidrule(lr){2-4}\cmidrule(lr){5-9}\cmidrule(l){10-13}
Shelf\&Tote \cite{zeng2017multi} & \cmark & \cmark & \xmark & 39 & 452 & 1 & $\cdot$ & 7K & 4.6 & - & - & bin, shelf \\
Rutgers APC \cite{rennie2016dataset} & \cmark & \cmark & \xmark & 25 & - & 1 & $\cdot$ & 10K & $<$5 & - & - & shelf \\
YCB-Video \cite{xiang2018posecnn} & \cmark & \cmark & \xmark & 21 & 92 & 1 & $\cdot$ & 130K & 4.5 & 86.1 & 47.3 & table \\
T-LESS \cite{hodan2017t} & \cmark & \cmark & \xmark & 30 & 20 & 3 & $\cdot$ & 48K & $\sim$7 & - & - & table \\
MP6D \cite{chen2022mp6d} & \cmark & \cmark & \xmark & 20 & 77 & 1 & $\cdot$ & 20K & 6.2 & - & - & table \\
PACE \cite{you2024pace} & \cmark & \cmark & \xmark & 238 & 300 & 1 & $\cdot$ & 55K & 4.7 & 85.5 & 41.5 & indoor, no shelf \\ \hline
OCID-Grasp \cite{suchi2019easylabel, ainetter2021end} & \cmark & \xmark & \ding{51}* & 89 & 96 & 1 & 75K & 2K & 7.5 & - & - & table, floor \\
MetaGraspNetv2 \cite{gilles2023metagraspnetv2} & \cmark & \cmark & \cmark & 82 & \underline{800} & 1 & - & 32K & 4.7 & 90.8 & 32.1 & bin \\
HouseCat6D \cite{jung2024housecat6d} & \cmark & \cmark & \cmark & \underline{194} & 41 & 2 & 10M & 25K & 6.7 & 96.0 & 23.5 & table \\
GraspNet-1B \cite{fang2020graspnet, fang2023robust} & \cmark & \cmark & \cmark & 88 & 190 & 2 & \underline{1.1B} & \textbf{97K} & \underline{8.9} & \underline{90.9} & \underline{35.2} & table \\
\rowcolor[gray]{0.9} \textbf{\textit{GraspClutter6D (Ours)}} & \cmark & \cmark & \cmark & \textbf{200} & \textbf{1,000} & \textbf{4} & \textbf{9.3B} & \underline{52K} & \textbf{14.1} & \textbf{77.1} & \textbf{62.6} & \textbf{bin, shelf, table} \\ \hline
\end{tabular}
}
\label{tab:1}
\end{table*}

\section{RELATED WORK}
This section reviews key methods and datasets in robotic perception and grasping. Table \ref{tab:1} summarizes real-world datasets for 6D object pose estimation and robotic grasping.

\textbf{Grasping in Cluttered Scenes.} Traditional parallel-jaw grasping \cite{glover2013bingham} relied on known object models, registering CAD models to scene point clouds for predefined grasps. Learning-based methods \cite{ten2017grasp, mahler2017dex, sundermeyer2021contact, fang2023anygrasp, ma2023towards, wu2024economic} subsequently enabled the grasping of unknown objects. GPD \cite{ten2017grasp} pioneered a deep network for 6-DoF grasps, while DexNet \cite{mahler2019learning} facilitated planar grasping through large-scale learning on synthetic data. Recent grasping networks, such as Contact-GraspNet \cite{sundermeyer2021contact} and AnyGrasp \cite{fang2023anygrasp}, directly predict dense 6-DoF grasps from visual scenes by learning from large-scale datasets \cite{eppner2021acronym, fang2020graspnet, fang2023robust}. Nevertheless, robust grasping in complex, cluttered environments remains challenging, as simulation datasets \cite{eppner2021acronym, gilles2023metagraspnetv2} suffer from sim-to-real transfer gaps, while existing real-world datasets \cite{fang2020graspnet, fang2023robust, jung2024housecat6d} primarily feature simplified, structured scenes with low occlusion and limited diversity. We address these limitations by providing an extensive real-world grasping dataset featuring highly cluttered scenes across diverse environments and objects.

\textbf{Segmentation and 6D Object Pose Datasets.}
Robotic perception primarily consists of segmentation \cite{xie2021unseen, back2022unseen} to locate object instances and 6D pose estimation \cite{he2021ffb6d, labbe2023megapose} for precise manipulation. Early segmentation datasets, such as OSD \cite{richtsfeld2012segmentation} and OCID \cite{suchi2019easylabel}, provided foundational RGB-D scenes but were limited in scale, offering at most 2K images. ArmBench \cite{mitash2023armbench} includes 53K warehouse images but lacks the 6D pose data critical for robust manipulation. For 6D object pose estimation, the BOP benchmark \cite{hodan2024bop} has advanced progress through standardized evaluations, featuring datasets such as YCB-Video \cite{xiang2018posecnn}, T-LESS \cite{hodan2017t}, and Rutgers APC \cite{rennie2016dataset}. However, these datasets contain relatively lightly cluttered setups—averaging 4.5 \cite{xiang2018posecnn} to 7 \cite{hodan2017t} objects per scene—with moderate occlusion (up to 47.3\% in \cite{xiang2018posecnn}). Similarly, the recent PACE dataset \cite{you2024pace}, despite encompassing 238 objects, maintains modest complexity with 4.7 objects per scene and 41.5\% occlusion. In contrast, GraspClutter6D offers 736K object poses in densely packed scenes (averaging 14.1 objects, 62.6\% occlusion) with 9.3B grasp annotations.

\textbf{Robotic Grasping Datasets.}
State-of-the-art robotic grasping methods rely on deep learning approaches, which necessitate large-scale, diverse datasets for robust performance. Early datasets such as Cornell \cite{lenz2015deep} and Jacquard \cite{depierre2018jacquard} provided valuable resources for grasp learning. However, they were limited to planar grasping in single-object scenes. OCID-Grasp \cite{ainetter2021end} extended this paradigm to cluttered scenes but offered only 75K planar grasp annotations. Synthetic datasets can be scalable alternatives: DexNet \cite{mahler2017dex} with 6.7M planar grasps, ACRONYM \cite{eppner2021acronym} with 6-DoF grasps for 8K objects, and MetaGraspNetv2 \cite{gilles2023metagraspnetv2} with 8,000 synthetic scenes. However, they face significant sim-to-real gaps, such as inaccurate sensor noise and contact dynamics. Among real-world 6-DoF grasp datasets, only a few public datasets exist. GraspNet-1B \cite{fang2020graspnet, fang2023robust}, a widely adopted grasping benchmark, provides 1.1B grasp poses across 190 scenes. Though it established a foundation for 6-DoF grasping research, it remains confined to simple green tabletop environments with relatively low complexity (35.2\% occlusion) and no background variation. The recent HouseCat6D dataset \cite{jung2024housecat6d} enhances diversity with 194 objects, yet remains limited in scale and variation, offering only 10M grasps for 41 table-only scenes with low occlusion (23.5\%). GraspClutter6D bridges these gaps by providing 9.3B 6-DoF grasp annotations across 1,000 highly cluttered scenes. % spanning diverse environments including bins, shelves, an.

\section{GraspClutter6D Dataset}

This section presents our systematic approach to establish a comprehensive dataset for robotic manipulation in complex environments. We detail the data acquisition and annotation process, followed by key statistics and quality evaluation.

\subsection{Data Acquisition}

\begin{figure}[t!]
\vspace*{0.3cm}
    \centering
         \includegraphics[width=0.9\columnwidth]{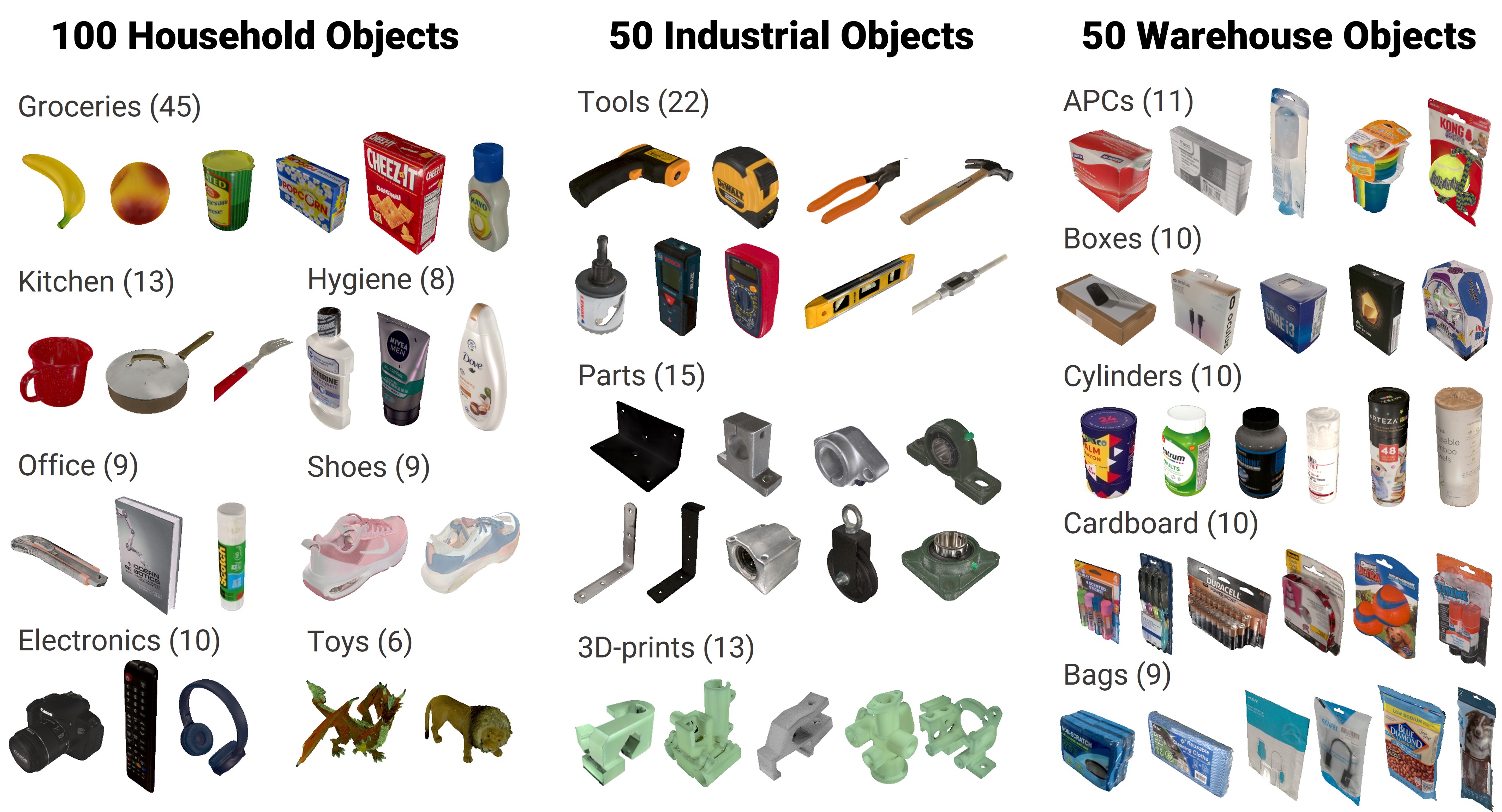}
            \caption{\textbf{Diverse 3D object models from GraspClutter6D}. The numbers in brackets denote objects per category.}
    \label{fig2}
\end{figure}

\begin{figure}[t!]
    \centering
         \includegraphics[width=0.9\columnwidth]{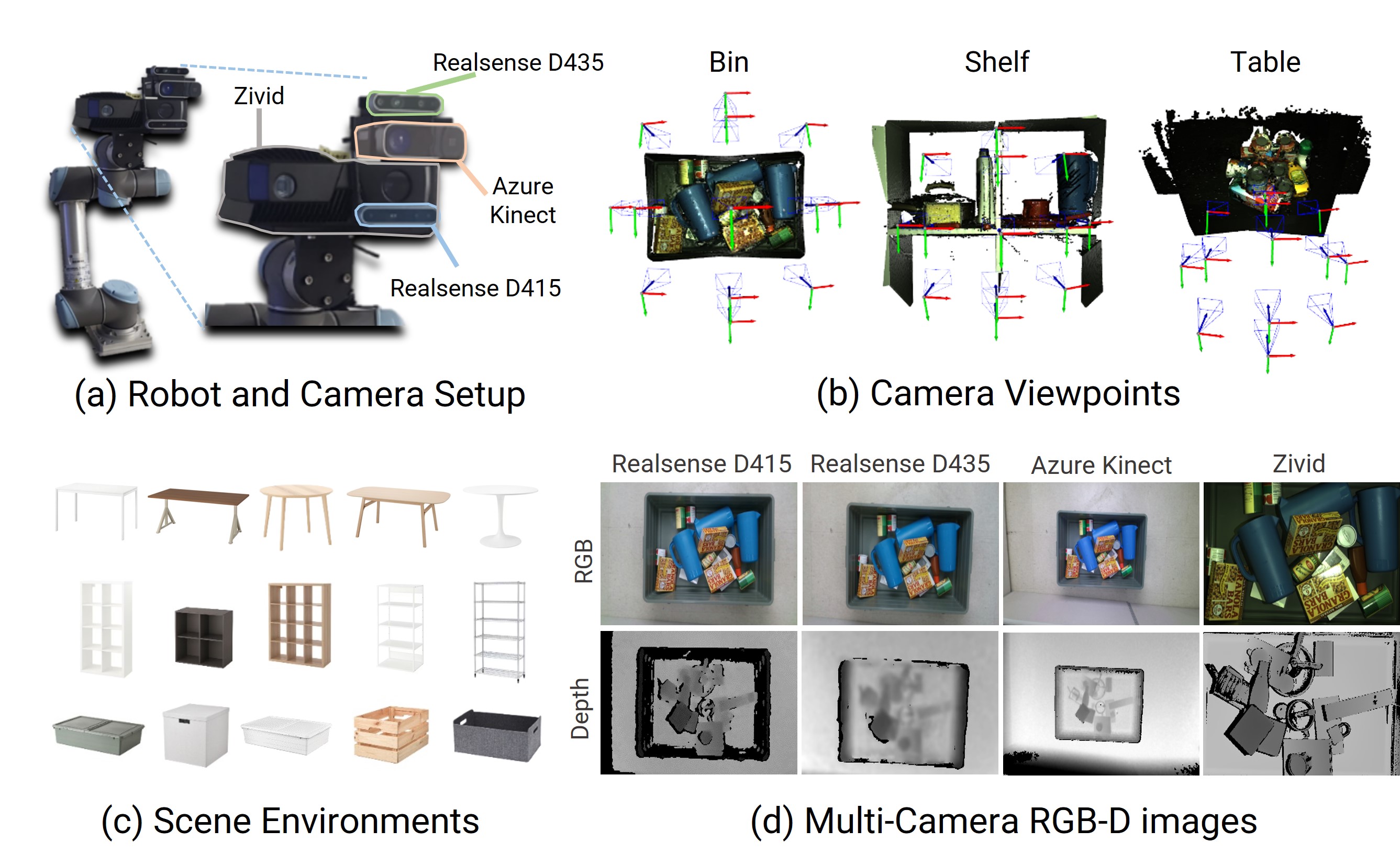}
    \caption{\textbf{Multi-camera capture system:} (a) a UR5 with four cameras, (b) multiple camera viewpoints, (c) bin, shelf, and table setup, (d) RGB-D image comparison.}
    \label{fig3}
\end{figure}

\textbf{3D Object Models.} Our dataset comprises 200 objects (Fig. \ref{fig2}), carefully curated to span household, warehouse, and industrial domains. We scanned 108 custom objects and incorporated 92 items from established benchmarks—YCB \cite{calli2015benchmarking}, HOPE \cite{tyree20226}, GraspNet-1B \cite{fang2020graspnet, fang2023robust}, DexNet \cite{mahler2017dex}, and APC \cite{correll2016analysis}—to ensure compatibility. High-quality 3D textured models were generated for custom objects using an Artec Leo scanner followed by post-processing by experts. This process yielded watertight, hole-free meshes with detailed textures. Reflective and transparent objects, which typically cause scanning artifacts, were coated with non-reflective gray spray to capture accurate geometry. The resulting 3D models are available in unified formats, supporting annotation and potential synthetic data generation. We also provide purchase links for all objects for future robotics research.

\textbf{Hardware Setup.} We developed a multi-sensor robotic capture system for efficient multi-viewpoint scene collection (Fig. \ref{fig3} (a)-(c)). Four RGB-D cameras were mounted on a UR5 robot arm using a rigid rig: three low-cost commercial sensors (RealSense D415, D435, and Azure Kinect) and one sub-millimeter high-precision sensor (Zivid One+ M). These cameras were synchronized to capture simultaneous data. To create diverse yet reproducible environments, we selected 5 IKEA furniture pieces from each of three categories (tables, shelves, and bins) with 5 background variations per category, yielding 75 unique configurations.

\textbf{Camera Calibration.} We performed extensive calibration for high-quality RGB-D data. We first conducted intrinsic calibration using a ChArUco board \cite{garrido2014automatic}. Then, relative poses between cameras were obtained using MC-Calib \cite{rameau2022mc}, followed by bundle adjustment with PnP-RANSAC \cite{lepetit2009ep} applied to matched points \cite{detone2018superpoint, sarlin2020superglue} on key frames. Camera poses relative to the robot were determined using ArUco markers \cite{garrido2014automatic}. To address systematic depth errors in low-cost RGB-D cameras, we applied depth correction through least-squares fitting with linear models \cite{hodan2017t}. This reduced mean absolute depth errors from \SI{10.2}{\mm} to \SI{5.0}{\mm} for the RealSense D415, from \SI{15.6}{\mm} to \SI{7.0}{\mm} for the RealSense D435, and from \SI{25.4}{\mm} to \SI{15.8}{\mm} for the Azure Kinect. Our dataset provides undistorted, depth-corrected RGB-D pairs with camera poses, requiring no additional post-processing.

\textbf{Scene Capture.} We generated 1,000 cluttered scenes by randomly placing 5-20 objects in each configuration. Each scene was recorded from 13 viewpoints (Fig. \ref{fig3} (b))—one centered and twelve peripheral angles—ensuring comprehensive visual coverage. This process yielded 52,000 RGB-D images across all cameras. Example RGB-D images from different cameras are shown in Fig. \ref{fig3} (d), capturing diverse fields of view, illuminations, and depth characteristics. Overall, the dataset features high scene complexity with an average of 14.1 object instances per image and 77.1\% visibility, providing multi-camera, multi-view real-world data for robotic manipulation in cluttered environments.

\begin{figure}[t!]
\vspace*{0.3cm}
    \centering
         \includegraphics[width=0.9\columnwidth]{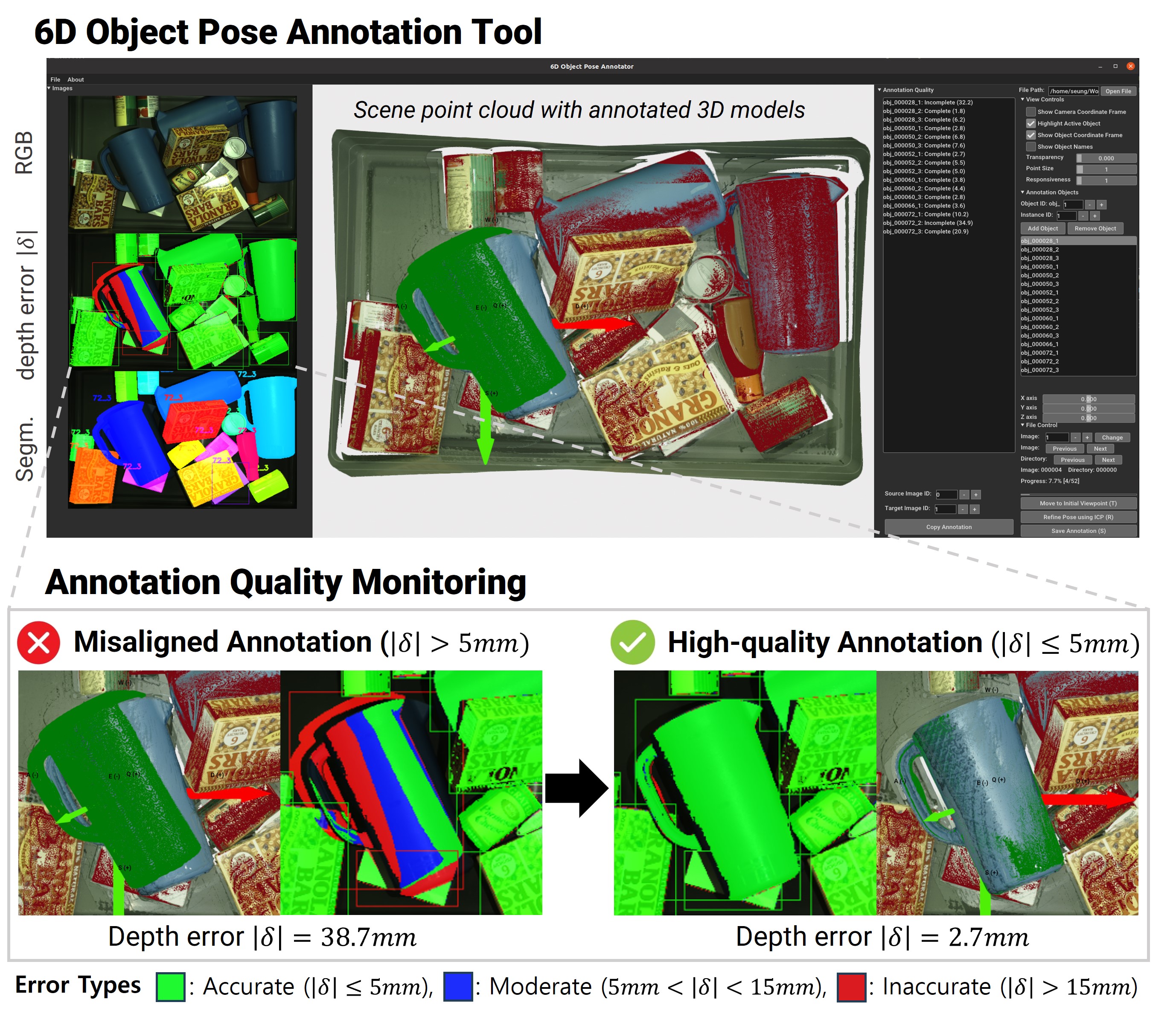}
    \caption{\textbf{Pose annotation tool with quality monitoring.} Top: The interface displays point cloud, RGB, and annotation quality. Bottom: Comparison between misaligned and high-quality annotation. Color-coded error visualization guides annotators to achieve the target accuracy threshold of \SI{5}{\mm}.}
    \label{fig4}
\end{figure}

\subsection{Annotation Pipelines}

\textbf{Object Pose Annotation.} GraspClutter6D is annotated through crowd-sourcing with a custom annotation tool (Fig. \ref{fig4}). To ensure high-quality and efficient annotation, we used two strategies: 1) a quality monitoring system providing immediate visualization of annotation errors, and 2) annotation propagation, wherein each scene is annotated once on an integrated point cloud and then propagated to all views. 

We first generated integrated point clouds of each scene by merging data from 13 different Zivid camera views. Annotators then aligned 3D object models to these sub-millimeter accurate point clouds, targeting a mean depth error below \SI{5}{\mm} per object. Our tool visualized depth errors between annotated object models and scene point clouds, allowing annotators to monitor quality and refine alignments through additional adjustment and ICP \cite{rusinkiewicz2001efficient} refinement upon saving each annotation. Independent experts verified all annotations through a double-review process. The verified poses were propagated to all camera views using camera intrinsic parameters and extrinsic poses. This process yielded $735,545$ annotated instances with 6D object poses and segmentation masks. The annotation tool is available on our project website.

\textbf{Grasp Pose Annotation.} We adopted a two-stage pipeline based on well-established methods in \cite{fang2020graspnet, fang2023robust} for 6-DoF grasp annotation: 1) object-level annotation with the force-closure metric, and 2) scene-level annotation with grasp projection and collision detection. We uniformly downsampled 200 object models in voxel space, sampling 14.4K grasps per point. For each grasp, the force closure metrics \cite{nguyen1988constructing, mahler2017dex} were calculated by varying the friction coefficients from 1.0 to 0.1. Then, object-level grasps were projected to scene-level grasps using annotated 6D object poses. Finally, collision checking was performed by filtering invalid grasps that overlap with the reconstructed scene point cloud. This process yielded 9.3B collision-free 6-DoF grasp poses for a parallel-jaw gripper (avg. 178K grasps per image), providing the largest grasp annotations in real cluttered scenes.

\begin{table}[h!]
\caption{\textbf{Comparison of 6D object pose annotation accuracy across datasets}, measured by depth differences ($\delta$) between captured and rendered depth at annotated poses. ($\mu_{|\delta|}$ and $med_{|\delta|}$: mean and median absolute depth differences, $\sigma_{\delta}$: standard deviation of depth differences in \SI{}{mm}.)}
\centering
\resizebox{0.9\columnwidth}{!}
{
\def\arraystretch{1.0}%  1 is the default, change whatever you need

\begin{tabular}{l|l||rrr}
\hline
Dataset & Sensor & \multicolumn{1}{c}{\textbf{$\mu_{|\delta|}$}} & \multicolumn{1}{c}{\textbf{$med_{|\delta|}$}} & \multicolumn{1}{c}{\textbf{$\sigma_{\delta}$}} \\ \hline
T-LESS \cite{hodan2017t} & Camarine & 4.28 & 2.46 & 7.72 \\
 & Kinect v2 & 8.40 & 5.45 & 11.36 \\ \hline
LINEMOD \cite{hinterstoisser2013model} & Kinect v2 & 5.89 & 5.57 & 1.47 \\ \hline
YCB-Video \cite{xiang2018posecnn} & Xtion Pro Live & 3.95 & 3.66 & 2.26 \\ \hline
MP6D \cite {chen2022mp6d} & Tuyang FM851-E2 & 3.54 & 2.70 & 0.17 \\ \hline
GraspNet-1B \cite{fang2020graspnet} & RealSense D435 & 7.69 & 4.95 & 14.30 \\
 & Azure Kinect & 14.79 & 9.54 & 20.20 \\ \hline
\textbf{GraspClutter6D} & Zivid & \textbf{3.22} & \textbf{1.55} & 11.10 \\
\textbf{(Ours)} & RealSense D415 & 5.71 & 3.77 & 14.67 \\
 & RealSense D435 & 7.02 & 4.58 & 13.82 \\
 & Azure Kinect & 13.85 & 6.83 & 32.59 \\ \hline
\end{tabular}

}
\label{tab:2}
\end{table}

\subsection{Dataset Statistics}
\textbf{Annotation Accuracy.}      
The annotation quality of 6D object poses was measured using the depth difference ($\delta = d_r - d_c$) as proposed by \cite{hodan2017t}. This metric calculates the difference between rendered depth ($d_r$) based on annotated object poses and captured depth ($d_c$) acquired from sensors for valid depth pixels (non-zero and finite) in both images. Values exceeding \SI{5}{\cm} were excluded following the standards to remove outliers from occlusion. Thus, $\delta$ quantifies pose annotation errors with sensor depth precision.

Table \ref{tab:2} presents the comparison of annotation accuracy with existing datasets. The mean and median absolute depth difference ($\mu_{|\delta|}$ and {$med_{|\delta|}$}) measure error magnitude, while standard deviation ($\sigma_{\delta}$) indicates consistency of annotation and sensor measurement. For a fair comparison with other datasets, the values of GraspClutter6D were calculated on opaque objects \cite{tyree20226}, as reflective and transparent objects in our dataset inherently produce measurement artifacts in depth sensors. The results show that GraspClutter6D achieved high annotation accuracy with $\mu_{|\delta|}$ of \SI{3.22}{\mm} for the Zivid sensor. Compared to GraspNet-1B \cite{fang2020graspnet, fang2023robust}, GraspClutter6D yielded higher accuracy with $\mu_{|\delta|}$ values of \SI{7.02}{\mm} and \SI{13.85}{\mm} for RealSense D435 and Azure Kinect, respectively, compared to their reported \SI{7.69}{\mm} and \SI{14.79}{\mm}. 

\begin{figure}[t!]
    % \vspace{0.3cm}
    \centering
         \includegraphics[width=0.95\columnwidth]{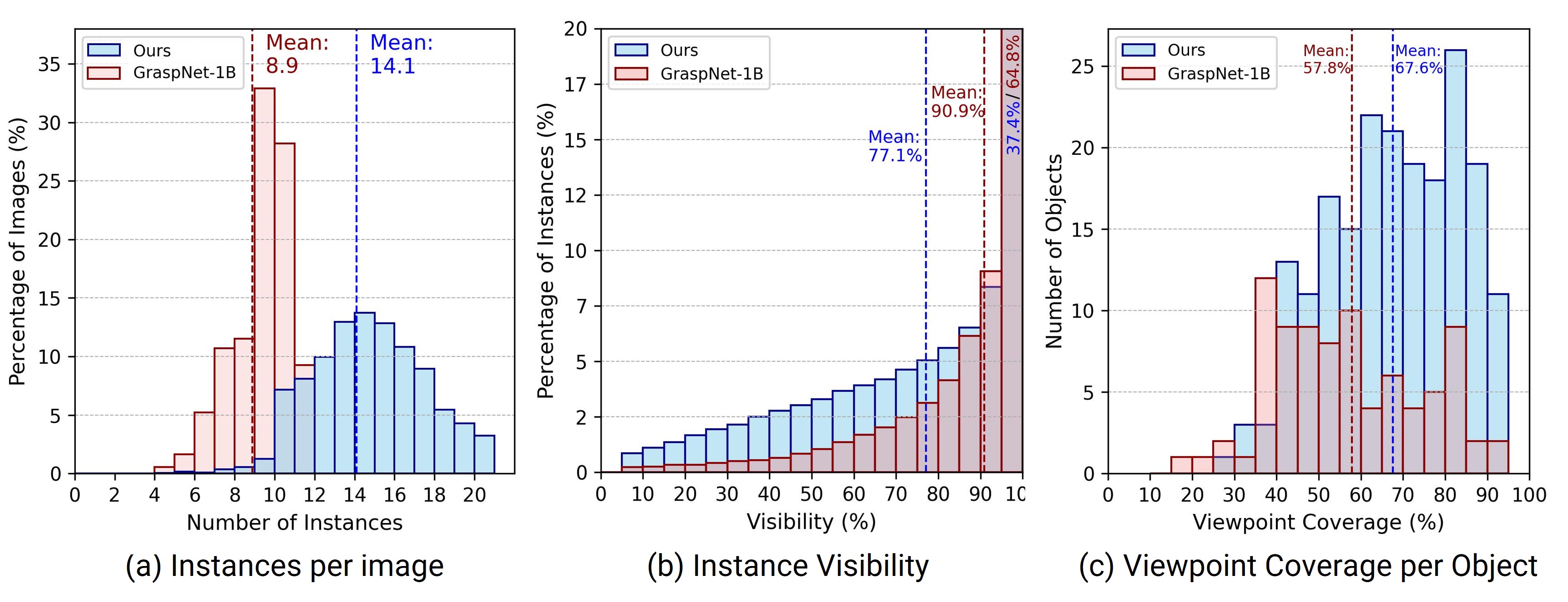}
    \caption{\textbf{GraspClutter6D statistics compared to GraspNet-1B}, showing comprehensive coverage of our dataset. 
    }
\label{fig6}
\end{figure}

\begin{figure}[t!]
    \centering
         \includegraphics[width=0.9\columnwidth]{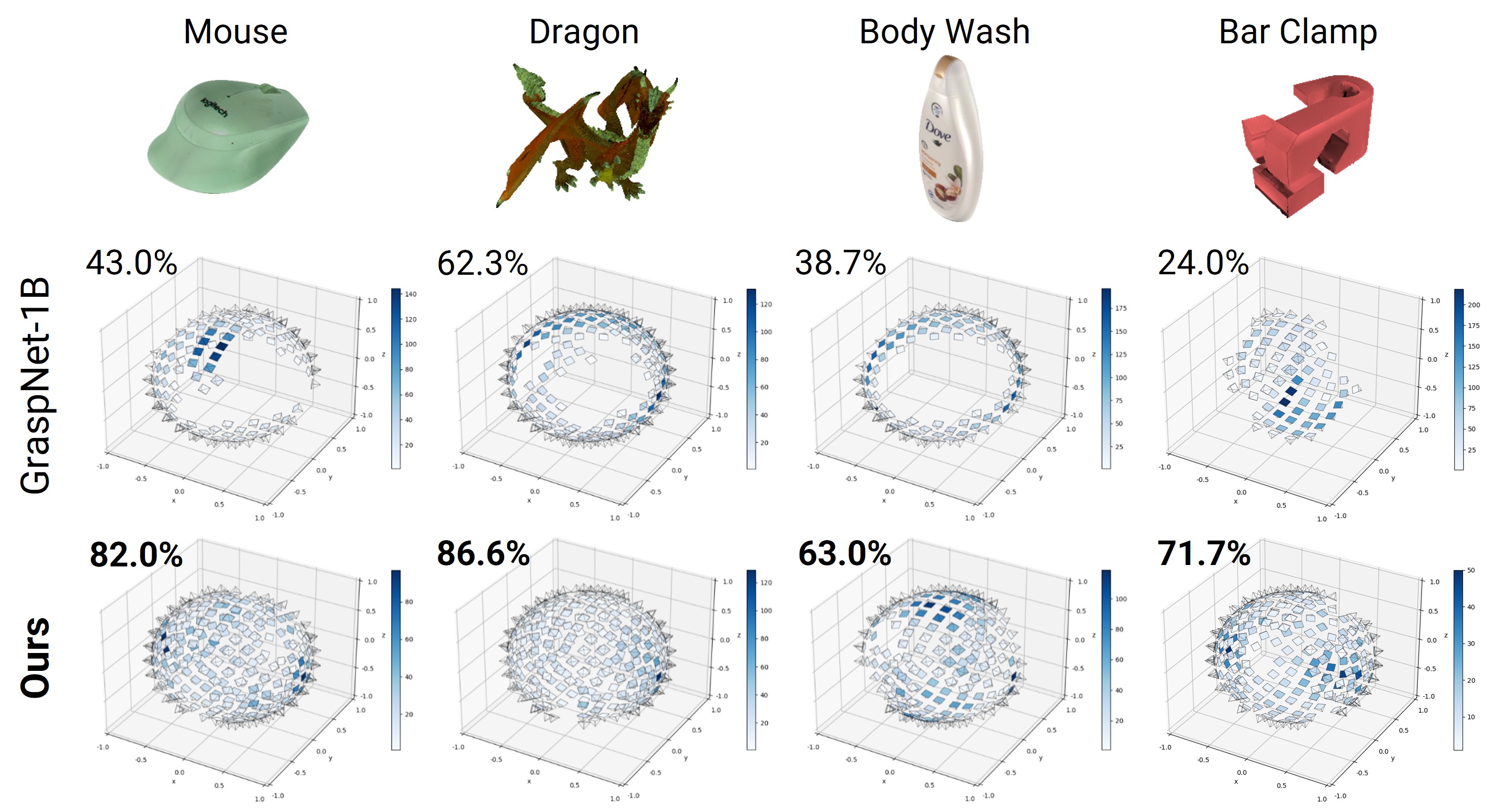}
    \caption{\textbf{Viewpoint coverage comparison} between GraspNet-1B (top) and GraspClutter6D (bottom). The heatmaps represent object-to-camera orientations discretized into 300 uniform spherical bins. Darker blue indicates high density.}
    \label{fig7}
\end{figure}

\textbf{Scene Statistics.} Our dataset comprises 1,000 real scenes featuring 200 objects with 735,545 instances annotated with 6D poses and segmentation masks, and 9.3 billion collision-free 6-DoF grasp poses. As shown in Fig. \ref{fig6} (a)-(c), our dataset includes highly cluttered scenes (avg. 14.1 instances and 77.1\% visibility), offering greater scene complexity than the GraspNet-1B (avg. $8.9$ instances with 90.9\% visibility). Object-to-camera distances range from \SI{0.31}{\m} to \SI{1.37}{\m} (average \SI{0.77}{\m}), with object diameters between \SI{4.5}{\cm} and \SI{47.3}{\cm} and weights from \SI{6}{\g} to \SI{2250}{\g} (mean \SI{199.5}{\g})—dimensions suitable for common robotic grippers. 

\textbf{Viewpoint Coverage.} Our dataset provides diverse viewpoint coverage through multi-view capture in bin, shelf, and table environments. We quantify this coverage by discretizing the object-to-camera rotations in spherical space into 300 uniform bins. As shown in Fig. \ref{fig6} (d), the mean viewpoint coverage for all objects in GraspClutter6D is $67.6$\%, higher than $57.8$\% for GraspNet-1B. We also compared the viewpoint coverages for the 32 objects that overlap between the two datasets in Fig. \ref{fig7}. GraspClutter6D achieves a mean viewpoint coverage of $73.8$\% for these overlapped objects, significantly higher than $48.5$\% of GraspNet-1B, due to its larger scale and more diverse capture environments.

\textbf{Dataset Splits.} For standardized evaluation, we provide two main splits: (1) A cross-object setup that tests generalization to novel objects. This split uses 68 unseen YCB-HOPE objects \cite{calli2015benchmarking, tyree20226} for testing (12K images, 235 scenes), widely used robotic research benchmark objects, and 132 objects for training (32K images, 413 scenes). (2) An intra-object setup focused on 21 common YCB-Video objects \cite{xiang2018posecnn} used in pose estimation benchmarks. This configuration includes 19K training images from 370 scenes and 4K test images from 74 scenes. The dataset also provides metadata for researchers to create custom splits for specific tasks.

\section{Experiments}

We conducted experiments with two primary objectives: (1) to evaluate the effectiveness of our dataset as a training resource for grasp detection and perception models, (2) to benchmark state-of-the-art methods on GraspClutter6D for future research. The following sections provide a comparative analysis of grasping performance across different training datasets, followed by benchmarks for segmentation, 6D object pose estimation, and 6-DoF grasp detection.

\subsection{Impact of GraspClutter6D on 6-DoF Grasp Detection}
\textbf{Experimental Design.} We investigated whether training on GraspClutter6D improves 6-DoF grasp detection performance. Using Contact-GraspNet \cite{sundermeyer2021contact} as our baseline network, we conducted comparative experiments in both simulated and real-world environments with three distinct datasets:
\begin{itemize}[leftmargin=10pt]
\item \textit{ACRONYM} \cite{eppner2021acronym}: A widely-used synthetic dataset with grasp annotations for 8,872 objects. Following protocols in \cite{sundermeyer2021contact}, we generated synthetic scenes by randomly placing 8-12 objects in stable, non-colliding poses on a table surface.
\item \textit{GraspNet-1B} \cite{fang2020graspnet, fang2023robust}: A popular real-world dataset whose training set contains 100 table scenes featuring 30 objects arranged in moderately cluttered configurations.
\item \textit{GraspClutter6D}: Our dataset comprising 413 highly cluttered scenes across table, shelf, and bin environments with 132 objects in the training set (cross-object setup).
\end{itemize}

We also benchmarked against \textit{AnyGrasp} \cite{fang2023anygrasp}, a state-of-the-art commercial grasp detection SDK trained on \textit{GraspNet-1B++}, a non-public dataset that extends GraspNet-1B with extra objects and scenes (144 objects across 268 real-world scenes). We used AnyGrasp with its default configuration. Contact-GraspNet was trained for 30K iterations with a batch size of 80 for each dataset.

\textbf{Simulation Setup.} We used PyBullet \cite{coumans2016pybullet} simulation from \cite{breyer2021volumetric} with two configurations: 1) \textit{packed} scenes, where 5 objects were placed in an upright pose without occlusion to test grasping in simple scenarios, and 2) \textit{pile} scenes with randomly dropped objects with increased complexity (5, 10, and 15 objects) to evaluate performance under varying occlusion levels. For each setup, we generated random scenes and executed the highest-confidence grasp using a Panda gripper. Rounds continued until all objects were cleared, no valid grasps remained, or two consecutive grasps failed. We conducted 500 simulation rounds per method and measured performance using standard metrics\cite{breyer2021volumetric}: Grasp Success Rate (GSR)—the ratio of successful lifts without slippage—and Declutter Rate (DR)—the average ratio of objects cleared per round. Objects out of workspace during manipulation were regarded as failures in DR. Single-view point clouds with Gaussian noise were used as input for all models.

\begin{table}[h!]
\caption{\textbf{Simulated grasping results.} We report grasping success rate (GSR) and declutter rates (DR).}
% \large
\centering
\resizebox{1.0\columnwidth}{!}
{
\def\arraystretch{1.1}%  1 is the default, change whatever you need
\setlength\tabcolsep{0.3em}

\begin{tabular}{l|l||cc|cccccc}
\hline
\multirow{3}{*}{Method} & \multirow{3}{*}{Train Data} & \multicolumn{2}{c|}{Packed} & \multicolumn{6}{c}{Pile} \\ \cline{3-10} 
 &  & \multicolumn{2}{c|}{{\ul 5 objects}} & \multicolumn{2}{c|}{{\ul 5 objects}} & \multicolumn{2}{c|}{{\ul 10 objects}} & \multicolumn{2}{c}{{\ul 15 objects}} \\
 &  & GSR & DR & GSR & \multicolumn{1}{c|}{DR} & GSR & \multicolumn{1}{c|}{DR} & GSR & DR \\ \hline \hline
AnyGrasp & GraspNet-1B++ \small\cite{fang2023anygrasp} & 77.5 & 78.9 & 58.6 & \multicolumn{1}{c|}{59.3} & 68.2 & \multicolumn{1}{c|}{53.0} & 71.6 & 49.5 \\ \hline
\multirow{3}{*}{\begin{tabular}[c]{@{}l@{}} Contact-\\GraspNet\end{tabular}} & ACRONYM \small\cite{eppner2021acronym} & 64.6 & 60.1 & 39.2 & \multicolumn{1}{c|}{28.8} & 44.8 & \multicolumn{1}{c|}{21.2} & 49.3 & 18.7 \\
 &  GraspNet-1B \small\cite{fang2023robust} & 74.5 & 65.2 & 62.7 & \multicolumn{1}{c|}{46.6} & 67.9 & \multicolumn{1}{c|}{36.3} & 71.0 & 33.4 \\
 & \textbf{GraspClutter6D} & \textbf{84.9} & \textbf{86.1} & \textbf{77.6} & \multicolumn{1}{c|}{\textbf{75.4}} & \textbf{77.2} & \multicolumn{1}{c|}{\textbf{64.7}} & \textbf{77.3} & \textbf{54.0} \\ \hline
\end{tabular}

}
\label{table:grasp_sim}
\end{table}

\begin{figure}[h!]
    \centering
         \includegraphics[width=\columnwidth]{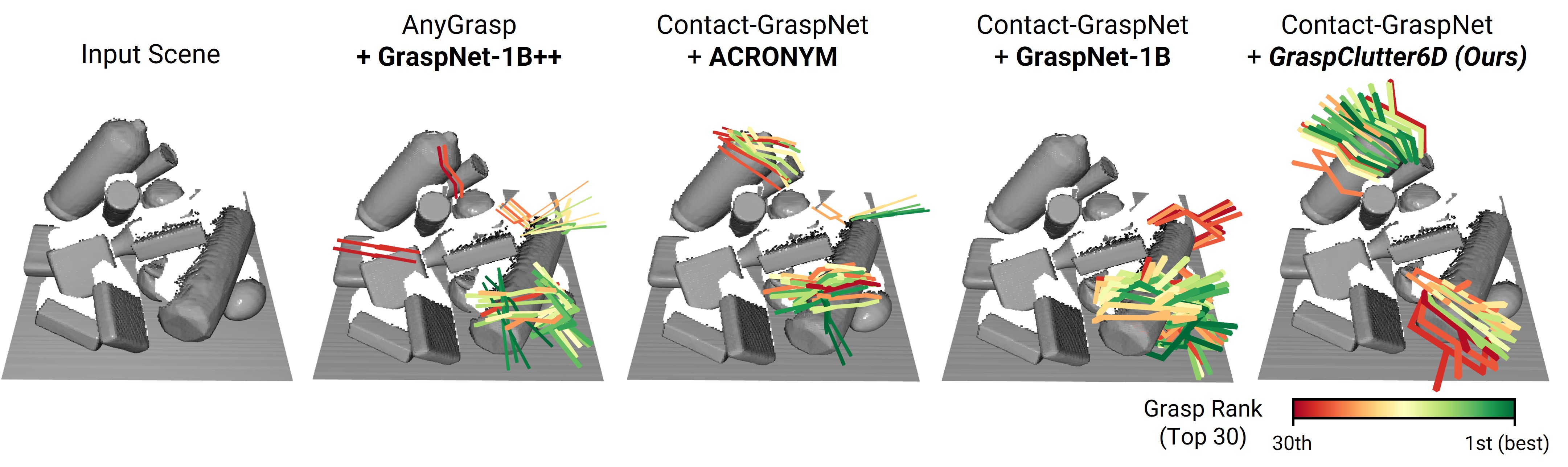}
    \caption{Grasp predictions (top 30) in a \textbf{simulated pile scene}.}
    \label{fig:grasp_sim}
\end{figure}

\textbf{Simulation Results.} Table \ref{table:grasp_sim} and Fig. \ref{fig:grasp_sim} compare grasping performance across different training datasets. The results demonstrate that Contact-GraspNet trained on our GraspClutter6D dataset significantly outperformed models trained on ACRONYM and GraspNet-1B across all test scenarios. Notably, our model achieved Grasp Success Rates (GSR) of 84.9\% in 5-object packed scenes and 77.3\% in more challenging 15-object pile scenarios, showing better performance than AnyGrasp. This suggests that the complexity and diversity of GraspClutter6D make it particularly effective for training robust grasping systems in cluttered environments.

\begin{figure}[h!]
    \centering
         \includegraphics[width=0.9\columnwidth]{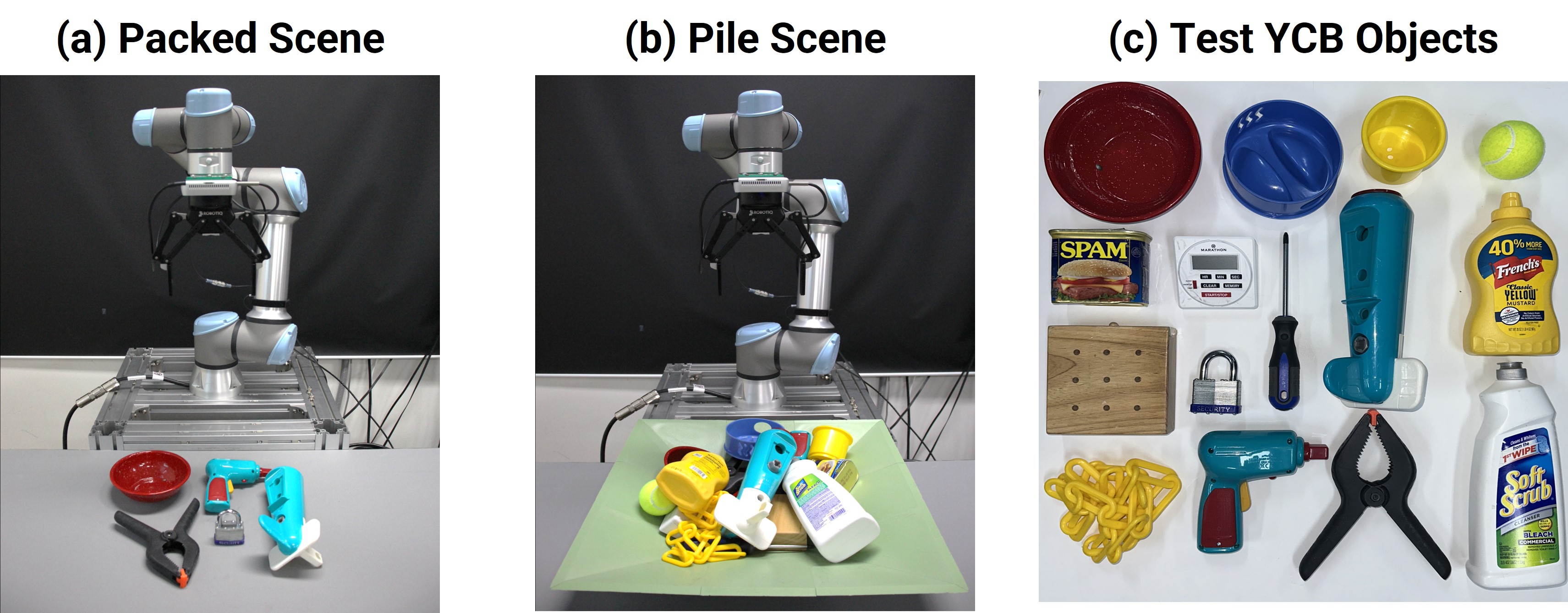}
    \caption{Real-world robotic grasping setup and test objects.}
    \label{fig:grasp_exp}
\end{figure}

\textbf{Real-world Setup.} We validated the effectiveness of GraspClutter6D in real-world robotic grasping. Similar to the simulation setup, we employed two configurations (Fig. \ref{fig:grasp_exp}): 1) \textit{packed} scenes with 5 objects placed upright on a table without occlusion, and 2) \textit{pile} scenes where objects were randomly poured into a bin, creating challenging environments at varying occlusion levels with 5, 10, and 15 objects. We used 15 unseen YCB objects \cite{calli2015benchmarking} as test items, ensuring no overlap with any training sets for fair assessment. For each attempt, models first predict grasps from the RealSense D435. Then we apply collision detection \cite{fang2020graspnet, fang2023robust} and background filtering (5cm threshold from segmented foreground objects using \cite{back2022unseen}). We executed the highest-confidence grasp among the remaining grasps using the Robotiq gripper on a UR5. For 5-object scenes, we standardized arrangements to minimize variance across methods, while for scenes with more than 5 objects, we used identical object sequences with random placement. We conducted 20 rounds for each method and setup, setting maximum consecutive failures at 3 ($\leq$10 objects) and 5 (15 objects) per round.

\begin{table}[h!]
\caption{\textbf{Real-world grasping results.} We report grasp success rate with successful grasps/total trials (in brackets).}
\large
\centering
\resizebox{1.0\columnwidth}{!}{
\def\arraystretch{1.15}%  1 is the default, change whatever you need
\setlength\tabcolsep{0.3em}
\begin{tabular}{l|l||c|ccc}
\hline
\multirow{2}{*}{Method} & \multirow{2}{*}{Train Data} & Packed & \multicolumn{3}{c}{Pile} \\ \cline{3-6} 
 &  & 5 objects & \multicolumn{1}{c|}{5 objects} & \multicolumn{1}{c|}{10 objects} & 15 objects \\ \hline \hline
AnyGrasp & GraspNet-1B++ \small\cite{fang2023anygrasp} & 62.7\small\hspace{0.1mm}(84/134) & \multicolumn{1}{c|}{66.4\hspace{0.1mm}\small(81/122)} & \multicolumn{1}{c|}{60.7\small\hspace{0.1mm}(131/216)} & 59.6\small\hspace{0.1mm}(252/423) \\ \hline
\multirow{3}{*}{\begin{tabular}[c]{@{}l@{}}Contact-\\ GraspNet\end{tabular}} & ACRONYM \small\cite{eppner2021acronym} & 32.1\small\hspace{0.1mm}(34/106) & \multicolumn{1}{c|}{27.8\small\hspace{0.1mm} (25/90)} & \multicolumn{1}{c|}{34.2\small\hspace{0.1mm} (42/123)} & 25.7 \small\hspace{0.1mm}(46/179) \\
 & GraspNet-1B \small\cite{fang2020graspnet, fang2023robust} & 77.5\small\hspace{0.1mm}(93/120) & \multicolumn{1}{c|}{62.5\small\hspace{0.1mm}(80/128)} & \multicolumn{1}{c|}{58.8\small\hspace{0.1mm}(124/211)} & 54.9\small\hspace{0.1mm}(230/419) \\
 & \textbf{GraspClutter6D} & \textbf{93.4}\small\hspace{0.1mm}(99/106) & \multicolumn{1}{c|}{\textbf{77.2}\small\hspace{0.1mm}(95/123)} & \multicolumn{1}{c|}{\textbf{74.0}\small\hspace{0.1mm}(174/235)} & \textbf{67.9}\small\hspace{0.1mm}(287/423) \\ \hline
\end{tabular}
}
\label{tab7}
\end{table}

\begin{figure}[h!]
    \centering
         \includegraphics[width=\columnwidth]{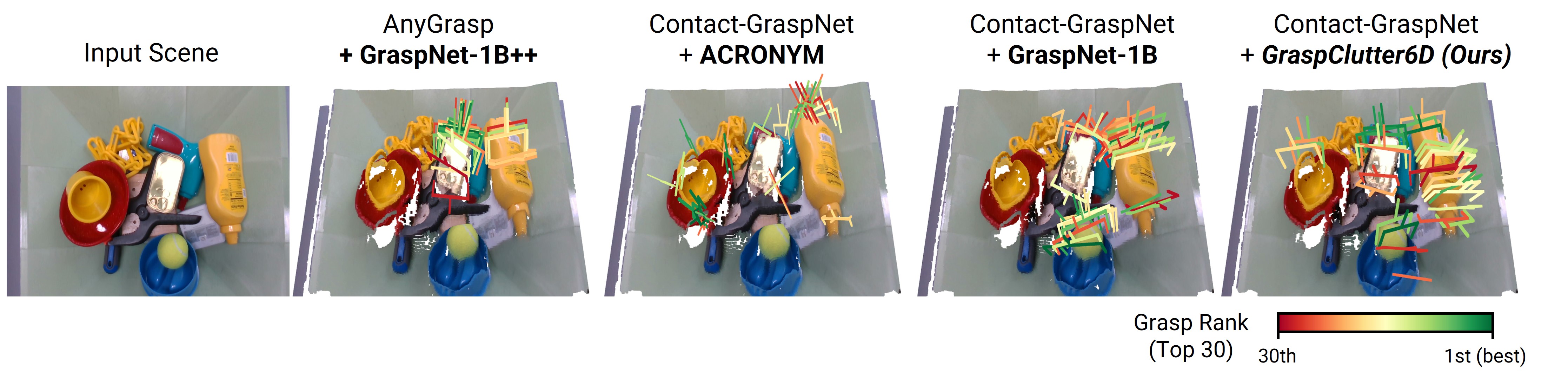}
    \caption{Grasp predictions (top 30) in a \textbf{real-world pile scene}.}
    \label{fig:grasp_real}
\end{figure}

\textbf{Real-world Results.} Table \ref{tab7} and Fig. \ref{fig:grasp_real} present a comparison of grasping performance in the real world. We observed that Contact-GraspNet trained on GraspClutter6D consistently outperformed other methods in all cases. Notably, the network trained on our dataset outperformed AnyGrasp in 15-object scenes (67.9\% and 59.6\%), highlighting GraspClutter6D as a valuable training source for grasping in complex cluttered environments. However, achieving robust grasping in highly cluttered scenes remains a significant challenge, offering an exciting direction for future research.

\subsection{Impact of GraspClutter6D on Robotic Perception}

\textbf{Setup.} We evaluated the effectiveness of GraspClutter6D as a training resource for perception tasks in comparison with GraspNet-1B. We used the standard YCB-Video \cite{xiang2018posecnn} dataset as our test set (21K images) and trained Mask2Former \cite{cheng2022masked} for segmentation and FFB6D \cite{he2021ffb6d} for 6D pose estimation using three different training configurations: (1) YCB-Video only (110K images), (2) YCB-Video + GraspNet-1B (56K additional images), and (3) YCB-Video + GraspClutter6D (14K additional images). We followed the standard evaluation protocols using COCO \cite{lin2014microsoft} metrics for segmentation, and ADD metrics \cite{xiang2018posecnn} for pose estimation.

\textbf{Results.} Tables \ref{table:segmeffect} and \ref{table:poseeffect} present segmentation and pose estimation performance on YCB-Video across different training configurations. Models trained with GraspClutter6D yield substantially greater performance improvements despite using 4$\times$ fewer additional training images (14K vs 56K). These results demonstrate that our dataset provides enhanced training benefits through increased complexity and diversity.

\begin{table}[h!]
\caption{\textbf{Segmentation} on YCB-Video (Mask2Former).}
\centering
\large
\resizebox{\columnwidth}{!}
{
\def\arraystretch{1.1}%  1 is the default, change whatever you need
\begin{tabular}{l||llll}
\hline
Train dataset & \multicolumn{1}{l}{AP} & \multicolumn{1}{l}{AP$_{50}$} & \multicolumn{1}{l}{AP$_{75}$} & \multicolumn{1}{l}{AR} \\ \hline \hline
YCB-Video \cite{xiang2018posecnn} & 57.4 & 73.1 & 66.4 & 60.8 \\
+ GraspNet-1B \cite{fang2020graspnet, fang2023robust} & 61.9 \small(+4.5) & 80.4 \small(+7.3) & 71.0 \small(+4.6) & 66.3 \small(+5.5) \\
\textbf{+ GraspClutter6D} & \textbf{66.1 \small(+8.7)} & \textbf{87.0 \small(+13.9)} & \textbf{74.7 \small(+8.3)} & \textbf{69.2 \small(+8.4)} \\ \hline
\end{tabular}

}
\label{table:segmeffect}
\end{table}

\begin{table}[h!]
\caption{\textbf{Pose estimation} on YCB-Video (FFB6D).}
\centering
\resizebox{0.9\columnwidth}{!}
{
\def\arraystretch{1.}%  1 is the default, change whatever you need

\begin{tabular}{l||ll}
\hline
Train dataset & ADD-S & ADD-S \\ \hline \hline
YCB-Video \cite{xiang2018posecnn} & 63.48 & 81.46 \\
+ GraspNet-1B \cite{fang2020graspnet, fang2023robust} & 64.24 \small (+0.76) & 85.09 \small(+3.63) \\
\textbf{+ GraspClutter6D} & \textbf{68.80 \small(+5.32)} & \textbf{85.76 \small(+4.30)} \\ \hline
\end{tabular}

}
\label{table:poseeffect}
\end{table}

\subsection{Instance Segmentation Benchmarks} 
\label{instsegbench}

\textbf{Setup.} We evaluated segmentation performance on unknown objects in cluttered scenes using state-of-the-art models: Mask R-CNN \cite{he2017mask}, Cascade Mask R-CNN \cite{cai2019cascade}, and Mask2Former \cite{cheng2022masked}. All models utilized a ResNet-50 FPN \cite{he2016deep, lin2017feature} backbone with the standard 1$\times$ training schedule and were trained and tested on GraspClutter6D using a cross-object setup, where training and test sets contain different objects. We also evaluated Grounded-SAM \cite{ren2024grounded}, a generalist foundation model trained on web-scale datasets, on GraspClutter6D without fine-tuning. For Grounded-SAM, we used a consistent prompt (``an object") and filtered out detections larger than half the image size to reduce false positives. We report average precision and recall using COCO metrics \cite{lin2014microsoft}.

\begin{table}[h!]
\caption{\textbf{Instance segmentation benchmarks.} $\ast$ denotes foundation models without domain-specific fine-tuning.}
\large
\centering
\resizebox{0.9\columnwidth}{!}
{
\def\arraystretch{1.0}%  1 is the default, change whatever you need
\setlength\tabcolsep{0.6em}

\begin{tabular}{l||cccc}
\hline
Method & AP & AP$_{50}$ & AP$_{75}$ & AR \\ \hline \hline
% SAM${\ast}$ \cite{kirillov2023segment} & 2.8 & 4.0 & 3.1 & \textbf{57.3} \\
% Detic${\ast}$ \cite{zhou2022detecting} & 5.5 & 9.5 & 5.5 & 47.7 \\
Grounded-SAM${\ast}$ \cite{ren2024grounded} & 16.2 & 22.3 & 17.9 & \textbf{55.3} \\ 
Mask R-CNN \cite{he2017mask} & 35.2 & 59.8 & 35.7 & 43.3 \\
Cascade R-CNN \cite{cai2019cascade} & 35.3 & 60.4 & 36.0 & 44.8 \\
Mask2Former \cite{cheng2022masked} & \textbf{43.5} & \textbf{69.0} & \textbf{44.9} & 52.2 \\ \hline
\end{tabular}
}
\label{table:segm}
\end{table}

\textbf{Results.} Table~\ref{table:segm} presents segmentation performance on GraspClutter6D. Mask2Former achieved the highest AP of 43.5, outperforming other models. Grounded-SAM demonstrated the highest recall of AR 55.3, but its low precision (AP of 16.2\%) reveals limitations in accurately delineating instance boundaries and distinguishing between individual objects in cluttered environments. These results suggest that domain-specific models remain more effective for segmentation in complex environments, though foundation models with adaptation strategies merit further exploration.

\subsection{6D Object Pose Estimation Benchmarks}

\textbf{Setup.} We evaluated object pose estimation on the standard 21 YCB-Video objects \cite{calli2015benchmarking, xiang2018posecnn} in GraspClutter6D using two types of approaches. First, models specialized for known objects—FFB6D \cite{he2021ffb6d} and GDR-Net \cite{wang2021gdr}—were trained on 19K images (intra-object setup) with their default configurations. Second, foundation models designed for novel object generalization—MegaPose \cite{labbe2023megapose} and FoundationPose \cite{wen2024foundationpose}—were evaluated, which were trained on large-scale synthetic datasets (2M and 1.2M images, respectively). We assessed performance using the ADD(-S) and the ADD-S metric \cite{hinterstoisser2012model, hodavn2016evaluation}. Following established protocols \cite{xiang2018posecnn}, we computed the area under the accuracy-threshold curve up to 10 cm. Results are categorized by occlusion levels based on instance visibility $v$: low ($v\geq0.9$), medium ($0.9>v\geq0.6$), and high occlusion ($0.6>v$). For fair comparison, all methods utilized 2D detections from Mask R-CNN \cite{he2017mask}, except FFB6D which jointly performs segmentation and pose estimation.

\begin{table}[h!]
\caption{\textbf{Pose estimation results.} ADD(-S)/ADD-S are reported across occlusion levels. $\ast$: foundation models.}
\large
\centering
\resizebox{\columnwidth}{!}
{
% \everymath{\medmuskip=1.5mu minus 1.5mu\thickmuskip=2mu minus 2mu}
\setlength\tabcolsep{0.3em}
\def\arraystretch{1.1}%  1 is the default, change whatever you need

\begin{tabular}{l||c|ccc}
\hline
{Method} & All & Low occ. & Medium occ. & High occ. \\ \hline \hline
FFB6D \cite{he2021ffb6d}& 36.2 / 59.3 & 45.1 / 69.5 & 35.1 / 59.5 & 13.2 / 26.8 \\
GDR-Net \cite{wang2021gdr}& 44.5 / 59.4 & 52.6 / 69.0 & 48.2 / 64.0 & 27.7 / 39.3 \\
MegaPose$\ast$ \cite{labbe2023megapose}& 69.6 / 78.6 & \textbf{77.9} / 86.7 & \textbf{71.5} / 80.5 & 42.3 / 51.7 \\
FoundationPose$\ast$ \cite{wen2024foundationpose}& \textbf{70.5} / \textbf{92.3} & 76.2 / \textbf{94.1} & 70.0 / \textbf{92.8} & \textbf{55.0} / \textbf{85.9} \\ \hline
\end{tabular}

}
\label{table:pose}
\end{table}

\textbf{Results.} Table \ref{table:pose} presents object pose estimation results on GraspClutter6D. Foundation models substantially outperformed domain-specific methods. These findings suggest that training on large-scale data is effective in improving generalizability for object pose estimation. All methods showed consistent performance degradation with increasing occlusion levels (e.g., MegaPose: 77.9 to 42.3 ADD(-S)), indicating that occlusion handling remains a persistent challenge.

\subsection{6-DoF Grasp Detection Benchmarks}
\textbf{Setup.} We evaluated four methods: Contact-GraspNet \cite{sundermeyer2021contact}, GraspNet-Baseline \cite{fang2020graspnet}, ScaleBalancedGrasp \cite{ma2023towards}, and EconomicGrasp \cite{wu2024economic}. All models were trained on GraspNet-1B \cite{fang2020graspnet, fang2023robust} with 100 train scenes and evaluated on (1) GraspNet-1B with 90 test scenes and (2) GraspClutter6D with 235 test scenes, cross-object setup. We compute the standard average precision (AP$_{\mu}$) \cite{fang2023robust}, which measures the average success rate of the top 50 predicted grasps at friction coefficient $\mu$ using the force-closure metric \cite{nguyen1988constructing, mahler2017dex}. We report \textbf{AP} as a primary metric, averaging AP$_\mu$ across friction coefficients from 0.2 to 1.2 at 0.2 intervals. A RealSense D435 camera was used in all datasets. To remove background-induced ambiguity, all models used workspace-cropped inputs and were evaluated for only foreground grasps ($\leq$5cm from target objects).

\begin{table}[h!]
\vspace{0.3cm}
\caption{\textbf{6-DoF grasp detection benchmark results.} (Train: GraspNet-1B, Test: GraspNet-1B and GraspClutter6D)}
\large
\centering
\resizebox{\columnwidth}{!}
{
\def\arraystretch{1.1}%  1 is the default, change whatever you need

\begin{tabular}{l||ccc|ccc}
\hline
\multirow{2}{*}{Method} & \multicolumn{3}{c|}{GraspNet-1B \cite{fang2023robust}} & \multicolumn{3}{c}{GraspClutter6D} \\ \cline{2-7} 
 & \textbf{AP} & AP$_{\textbf{0.8}}$ & AP$_{\textbf{0.4}}$ & \textbf{AP} & AP$_{\textbf{0.8}}$ & AP$_{\textbf{0.4}}$ \\ \hline \hline
Contact-GraspNet \cite{sundermeyer2021contact} & 19.07 & 24.34 & 11.05 & 10.73 & 13.18 & 5.75 \\
GraspNet-Baseline \cite{fang2020graspnet} & 23.43 & 27.81 & 19.56 & 17.95 & 21.54 & 13.53 \\
ScaleBalancedGrasp \cite{ma2023towards} & 44.85 & 53.31 & 39.31 & \textbf{22.69} & \textbf{27.57} & \textbf{16.95} \\
EconomicGrasp \cite{wu2024economic} & \textbf{51.63} & \textbf{61.55} & \textbf{43.72} & 21.67 & 26.62 & 15.90 \\ \hline
\end{tabular}

}
\label{table:graspbench}
\end{table}

\textbf{Results.} Table \ref{table:graspbench} compares the performance of grasping methods across datasets. All methods demonstrate significant performance degradation on GraspClutter6D compared to GraspNet-1B, with EconomicGrasp exhibiting a 29.96 AP reduction. This performance gap indicates that GraspClutter6D can serve as a challenging grasping benchmark, highlighting substantial room for improvement in real-world clutter.

\section{CONCLUSION}
We presented GraspClutter6D, a dataset for robotic grasping and perception in real-world cluttered environments. Our benchmark evaluations revealed a significant gap, as existing state-of-the-art methods show degraded performance in our high-complexity scenes. We demonstrated that GraspClutter6D serves as an effective training resource; models trained on our dataset achieved substantially improved grasping and perception, highlighting its value in developing robust systems to close this gap. For future work, we will focus on gripper-aware and reactive grasping motions, as we observed that simple approach trajectories often cause failures despite accurate grasp pose detection. By publicly releasing our dataset and toolkits, we hope this dataset will serve as a foundation for advancing robotic manipulation in cluttered environments.

% \section*{ACKNOWLEDGMENT}
% \begin{spacing}{0.7}
% {\scriptsize 
% This work was supported by the Technology Innovation Program(00442029, Development of Tactile Intelligence in Robotic Hands Based on Tactile Data Learning to Manipulate Irregular Multiple Types of Objects) funded by the Ministry of Trade Industry \& Energy(MOTIE, Korea)
% }
% \end{spacing}

\FloatBarrier
\bibliographystyle{IEEEtran}
\bibliography{IEEEabrv, references.bib}

\end{document}